\documentclass{article}
\usepackage{tech2025}

\usepackage{microtype}
\usepackage{graphicx}
\usepackage{booktabs} 

\usepackage{hyperref}

\usepackage[utf8]{inputenc} 
\usepackage[T1]{fontenc}    
\usepackage{url}            
\usepackage{amsfonts}       
\usepackage{nicefrac}       
\usepackage[table]{xcolor}         
\usepackage{colortbl}
\usepackage{amsmath}
\usepackage{amssymb}
\usepackage{mathtools}
\usepackage{amsthm}
\usepackage{mathrsfs}
\usepackage{xspace}
\usepackage{academicons}

\usepackage{booktabs}
\usepackage{adjustbox}
\usepackage{algorithm}
\usepackage{algorithmicx}
\usepackage{amsmath,amsfonts}
\usepackage{bbm}
\usepackage{subcaption}  

\usepackage{pifont}

\usepackage{colortbl}
\usepackage{multirow}

\newcommand{\Imagereward}{ImageReward}
\newcommand{\ImagerewardCite}{ImageReward~\cite{xu2023imagereward}}
\newcommand{\Hps}{HPSv2}
\newcommand{\HpsCite}{HPSv2~\cite{wu2023human}}
\newcommand{\Geneval}{GenEval}
\newcommand{\GenevalCite}{GenEval~\cite{ghosh2023geneval}}
\newcommand{\Dpg}{DPG-Bench}
\newcommand{\DpgCite}{DPG-Bench~\cite{hu2024ella}}
\newcommand{\Mps}{MPS}
\newcommand{\MpsCite}{MPS~\cite{MPS}}
\newcommand{\Clipscore}{Clip-Score}
\newcommand{\ClipscoreCite}{Clip-Score~\cite{ilharco_gabriel_2021_5143773}}

\usepackage[capitalize,noabbrev]{cleveref}

\theoremstyle{plain}

\theoremstyle{definition}

\theoremstyle{remark}

\definecolor{bigaired}{RGB}{156, 0, 0}
\definecolor{uclablue}{RGB}{39, 116, 174}
\definecolor{thupurple}{RGB}{102, 8, 116}
\definecolor{pkured}{RGB}{139, 0, 18}
\definecolor{panton}{RGB}{217, 51, 121}

\definecolor{darkred}{RGB}{200, 0, 0}
\definecolor{darkblue}{RGB}{0, 0, 200}
\definecolor{blue}{RGB}{0, 0, 250}

\definecolor{light}{RGB}{225, 250, 250}
\definecolor{lightgray}{RGB}{0.9, 0.9, 0.9}
\definecolor{lightred}{RGB}{250, 200, 200}
\definecolor{lightblue}{RGB}{210, 220, 250}
\definecolor{lightpurple}{RGB}{218,210,255}

\definecolor{doderblue}{RGB}{30, 144, 255}
\definecolor{select}{RGB}{222, 235, 247}
\definecolor{unselect}{RGB}{247, 207, 206}

\definecolor{myLinkColor}{HTML}{7D5BA6}     
\definecolor{myCiteColor}{HTML}{9A4D92}     
\definecolor{myURLColor}{HTML}{5B7DB1}      
\hypersetup{
  colorlinks=true,
  linkcolor=myLinkColor,
  citecolor=myCiteColor,
  urlcolor=myURLColor
}

\usepackage[textsize=tiny]{todonotes}

\usepackage[most]{tcolorbox}

\usepackage{listings}

\usepackage{fontawesome5}  
\usepackage{listings}      
\usepackage[misc]{ifsym}
\usepackage{wrapfig}    
\usepackage[T1]{fontenc}

\usepackage{tcolorbox}
\usepackage{relsize}

\usepackage{adjustbox}
\usepackage{fancyhdr}
\usepackage{lipsum}
\usepackage{newtxtext}
\usepackage{wrapfig}
\usepackage{enumitem}
\definecolor{azblue}{RGB}{27,117,187}      

\usepackage[noend]{algpseudocode}   

\definecolor{bestcol}{RGB}{  0,102,204} 
\definecolor{goodcol}{RGB}{ 34,139, 34} 
\definecolor{deltaBg}{RGB}{220,230,255} 
\newcommand{\cmark}{\ding{51}}  

\usepackage{lmodern}  

\usepackage{tikz}
\usetikzlibrary{tikzmark,decorations.pathreplacing,calc}
\usepackage{stackengine}
\stackMath
\definecolor{lightgreen}{RGB}{0,150,0}  

\usepackage{titletoc}

\newtheoremstyle{rqstyle}%
  {\topsep}            
  {\topsep}            
  {}                   
  {}                   
  {\bfseries}    
  {:}                  
  {.5em}               
  {}                   

\theoremstyle{rqstyle}

\crefname{researchquestion}{Research Question}{Research Questions}

\definecolor{propose}{HTML}{EF8E8D}
\definecolor{solve}{HTML}{5755A3}

\definecolor{humanred}{RGB}{180, 50, 50}
\definecolor{envgreen}{RGB}{50, 140, 80}

\definecolor{paleviolet}{HTML}{E1EEFC}
\definecolor{lightgrey}{RGB}{247, 247, 247}
\newenvironment{leapabstract}{
  \begin{tcolorbox}[
    colback=lightgrey,
    colframe=white,
    boxrule=0pt,
    arc=10pt,
    left=16pt,
    right=16pt,
    top=12pt,
    bottom=12pt,
    width=\textwidth,
    enlarge left by=0mm,
    before skip=10pt,
    after skip=10pt
  ]
  \normalsize
}{
  \end{tcolorbox}
}

\makeatletter
\DeclareRobustCommand\onedot{\futurelet\@let@token\@onedot}
\def\@onedot{\ifx\@let@token.\else.\null\fi\xspace}
 
\def\ie{\emph{i.e}\onedot}

\makeatother

\begin{document}

\makeatletter
\def\icmldate#1{\gdef\@icmldate{#1}}
\icmldate{\today}
\makeatother

\makeatletter
\fancypagestyle{fancytitlepage}{
  \fancyhead{}  
  \lhead{\includegraphics[height=1.5cm]{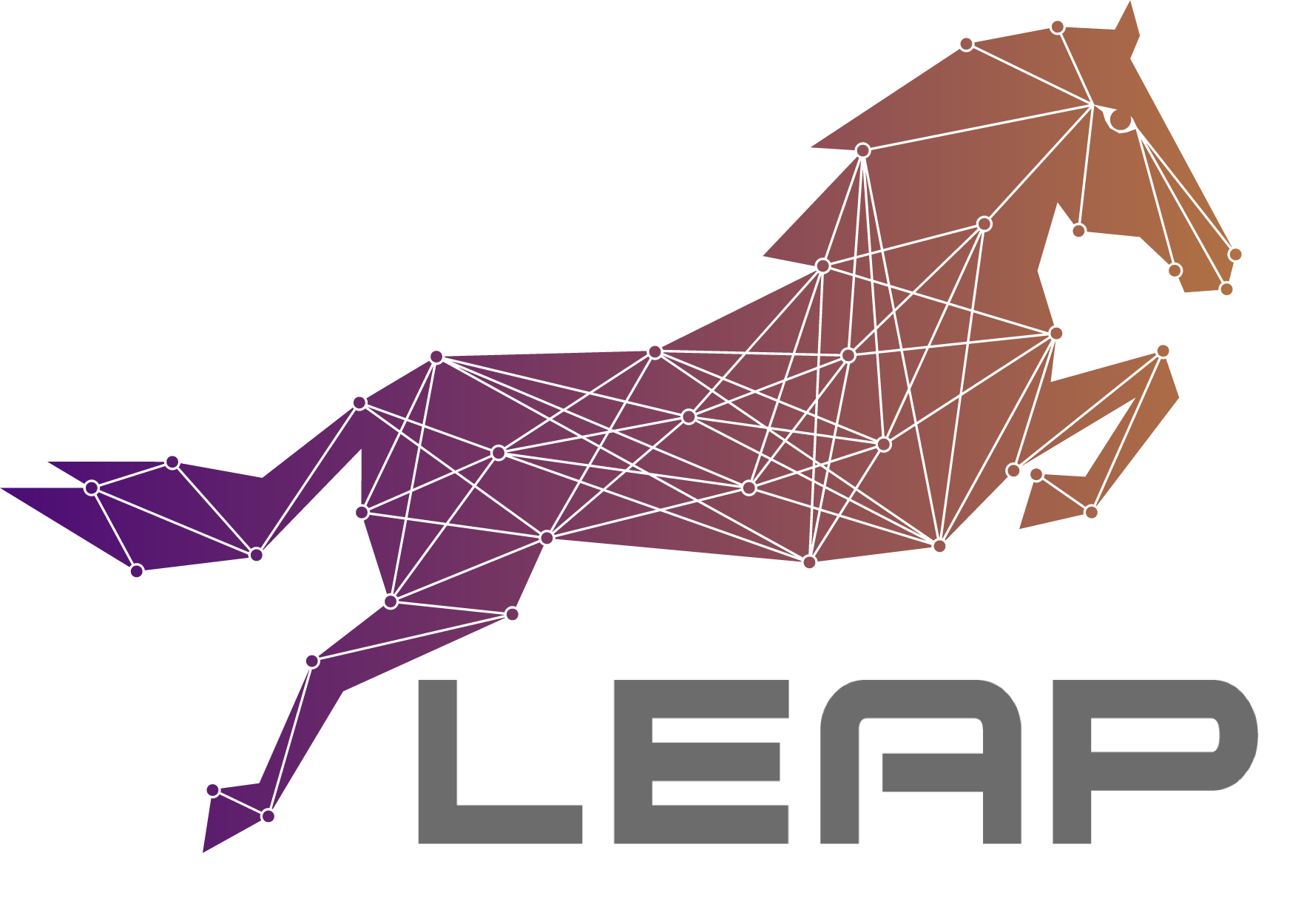}\hspace{5mm}}
  \rhead{\it \@icmldate}
  \cfoot{}
}
\makeatother



\icmltitle{Co-GRPO: Co-Optimized Group Relative Policy Optimization for Masked Diffusion Model}


\begin{icmlauthorlist}

Renping Zhou$^{\ast,1,2}$, \quad  Zanlin Ni$^{\ast,1}$,  \quad  Tianyi Chen$^{1}$, \quad  Zeyu Liu$^{1}$, \quad  Yang Yue$^{1}$, \\
\quad  Yulin Wang$^{1}$, \quad Yuxuan Wang$^{1}$, \quad  Jingshu Liu$^{1}$, \quad Gao Huang$^{\textrm{\Letter},1}$
\end{icmlauthorlist}

$^{1}$Leap Lab, Tsinghua University
$^{2}$Anyverse Dynamics
\icmlcorrespondingauthor{Gao Huang}{gaohuang@tsinghua.edu.cn}


\vskip 0.1cm

\printNotice{} 

\begin{leapabstract}
    
    
        Recently, Masked Diffusion Models (MDMs) have shown promising potential across vision, language, and cross-modal generation. However, a notable discrepancy exists between their training and inference procedures. In particular, MDM inference is a multi-step, iterative process governed not only by the model itself but also by various schedules that dictate the token-decoding trajectory (e.g., how many tokens to decode at each step). In contrast, MDMs are typically trained using a simplified, single-step BERT-style objective that masks a subset of tokens and predicts all of them simultaneously. This step-level simplification fundamentally disconnects the training paradigm from the trajectory-level nature of inference, leaving the inference schedules never optimized during training. In this paper, we introduce Co-GRPO, which reformulates MDM generation as a unified Markov Decision Process (MDP) that jointly incorporates both the model and the inference schedule. By applying Group Relative Policy Optimization at the trajectory level, Co-GRPO cooperatively optimizes model parameters and schedule parameters under a shared reward, without requiring costly backpropagation through the multi-step generation process. This holistic optimization aligns training with inference more thoroughly and substantially improves generation quality. Empirical results across four benchmarks—ImageReward, HPS, GenEval, and DPG-Bench—demonstrate the effectiveness of our approach. For more details, please refer to our project page: \href{https://co-grpo.github.io/}{https://co-grpo.github.io}.
    \vspace{10pt}
        
    \vskip -2cm
    \end{leapabstract}
    
\begin{figure}[h]
\centering
\includegraphics[width=1\textwidth]{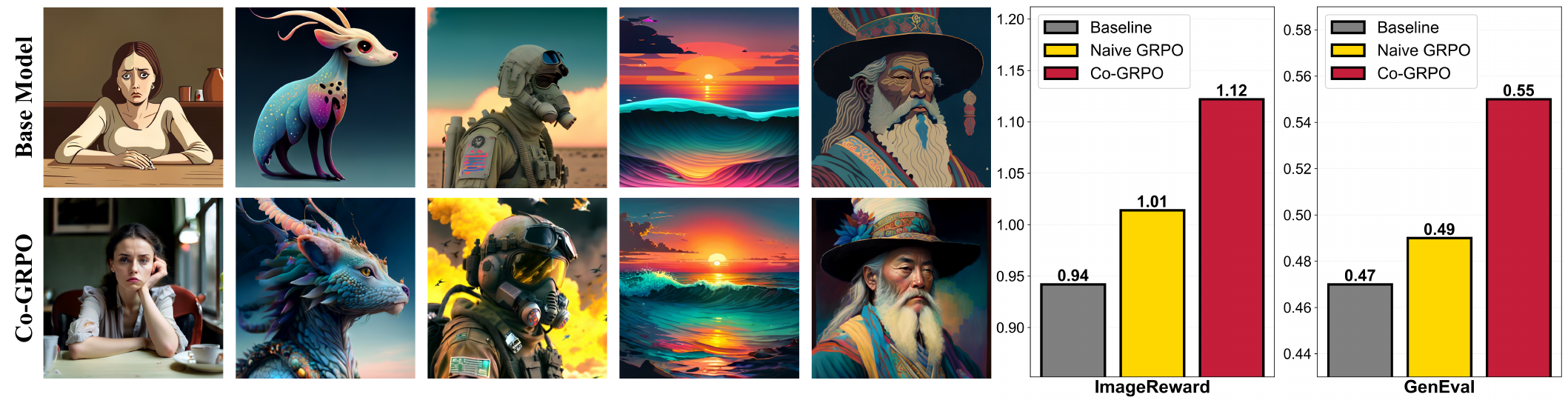}
\caption{\textbf{Qualitative and quantitative comparison of Co-GRPO against baseline approaches.} Through cooperative optimization of the MDM model and inference schedule, Co-GRPO produces images with markedly superior quality compared to baseline.
 Detailed prompts are provided in \cref{supptab:teaser-prompt}.}
\end{figure}
\section{Introduction}
\label{sec:intro}

Masked Diffusion Models (MDMs) have recently demonstrated remarkable success across diverse domains, including vision, language, and cross-modal applications, owing to their general-purpose modeling capabilities and significant potential for efficiency~\cite{nie2025large, zhu2025llada, ye2025dream,yang2025mmada,liang2025discrete}.

In the visual domain, MDMs offer an efficient alternative to autoregressive (AR) models. They preserve the benefits of unified token-based multi-modal modeling~\cite{bai2024meissonic,xie2024show,you2025llada} while delivering markedly higher generation efficiency: AR models often require hundreds of sequential steps, whereas many MDMs achieve decent-quality image synthesis in only 8–16 iterations~\cite{chang2022maskgit,chang2023muse,ni2024autonat,ni2024adanat}. 
This efficiency is achieved by their parallel decoding mechanism, where generation begins from a fully masked canvas and swiftly infills multiple tokens per step. 
To enable this capability, MDMs adopt a BERT-style training objective~\cite{devlin2019bert} that predicts all masked tokens conditioned on the visible context.

\begin{wraptable}{r}{0.48\columnwidth} 
    \footnotesize
    \vspace{-10pt}
        \begin{tabular}{l l}
        \toprule
            Inference Schedule  & Predefined Function \\
            \midrule
             re-mask ratio $r{(t)}$ & $r(t) = \cos{(\frac{t+1}{T}\cdot\frac{\pi}{2}})$ \\
             sample temperature $\tau_s{(t)}$ & $\tau_s{(t)}=1.0$ \\
             re-mask  temperature $\tau_r{(t)}$ & $\tau_r{(t)}=2\frac{T-t}{T}$ \\
             classifier-free guidance scale $s{(t)}$ & $s{(t)} = 9.0$ \\
        \bottomrule
        \end{tabular}
        \vspace{-7pt}
        \captionof{table}{An example of the manually designed inference schedule from Meissonic~\cite{bai2024meissonic}.}
        \vspace{-10pt}
        \label{tab:schedule-policy-example}
\end{wraptable}

However, a fundamental discrepancy exists between the training objective and the practical inference process. Iterative generation requires a set of \emph{inference schedules} governing the number of tokens decoded each step as well as other procedural parameters (see \cref{tab:schedule-policy-example}). These scheduling decisions, despite critically affecting final generation quality (see \cref{tab:simple-exp}), are never explicitly optimized during training. Specifically, the conventional BERT-style objective simplifies training to a single-step prediction problem: masking some tokens and decode all of them at once. This \emph{step-level} simplification circumvents the need for expensive backpropagation through the multi-step generation process but also inherently precludes the model from learning the \emph{trajectory-level} inference schedule. 
As a result, the model and the inference schedule, which jointly determine the final generation quality and should ideally be optimized together, end up being separated.
Practitioners must then rely on post-hoc, manually designed scheduling rules for inference, as shown in \cref{tab:schedule-policy-example}.

\begin{wrapfigure}{r}{0.48\textwidth}  
    \centering
    \footnotesize
    \vspace{-10pt}  
    \includegraphics[width=0.48\textwidth]{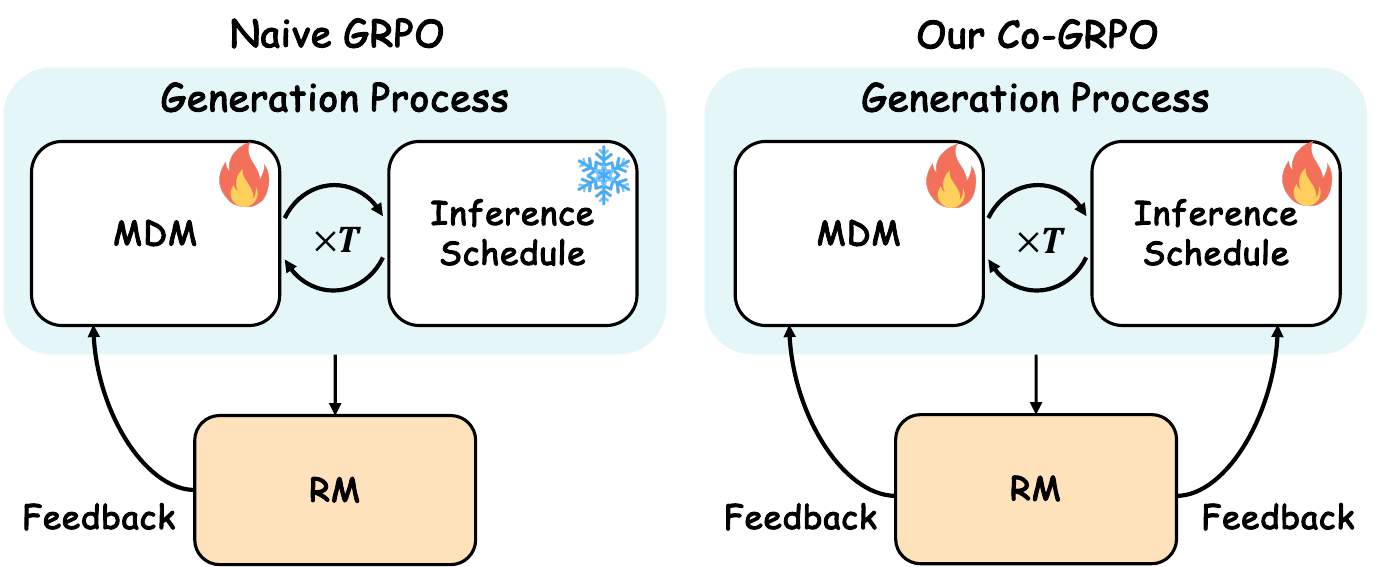}
    \caption{\textbf{Comparison between the conventional MDM post-training framework and our Co-GRPO.} Naive GRPO collects trajectories using a trainable MDM model under a fixed, predefined inference schedule. Our proposed Co-GRPO challenges this convention by cooperatively optimizing both the MDM model and the inference schedule based on the reward feedback.}
    \label{fig:motivation}
    \vspace{-10pt}  
\end{wrapfigure}

To address this issue, we introduce \textbf{Co-GRPO} (\textbf{C}o-\textbf{O}ptimized \textbf{G}roup \textbf{R}elative \textbf{P}olicy \textbf{O}ptimization).
We formulate a new  Markov Decision Process (MDP) that unifies the model and the inference schedule within a single GRPO-style policy. Leveraging the trajectory-level nature of GRPO, Co-GRPO is able to jointly optimize both components without the prohibitive cost of backpropagating through multi-step generation.
This holistic view stands in sharp contrast to Naive GRPO~\cite{luo2025maskgrpo,yang2025mmada}, which inherits the conventional separation between model and schedule and optimizes only the model parameters during training.
As conceptually illustrated in \cref{fig:motivation}, Co-GRPO instead treats model and schedule as cooperating policies driven by the same reward signal.
By aligning both components to a shared objective, Co-GRPO optimizes the entire generation trajectory instead of focusing solely on model parameters, leading to significantly better performance.

Our approach achieves substantial improvements in visual quality and reward alignment for MDMs by cooperatively optimizing both the model and inference schedule, producing outputs that exhibit both superior aesthetics and enhanced prompt adherence. We demonstrate significant performance gains across four diverse text-to-image benchmarks. On reward model-based benchmarks, Co-GRPO substantially surpasses the Naive GRPO baseline method, improving \Imagereward \ score from 0.942 to \textbf{1.122} and \Hps \ score from 28.83 to \textbf{29.37}. Moreover, the cooperatively optimized policy demonstrates strong generalization capability, achieving significant zero-shot improvements on both GenEval and DPG-Bench benchmarks without requiring additional fine-tuning.

Our contributions are summarized as follows:
\begin{itemize}
    \item We identify and formalize the fundamental mismatch between the step-level BERT-style training objective and the trajectory-level inference process in MDMs, revealing that the inference schedule—despite critically affecting generation quality—remains decoupled from training and thus unoptimized in existing approaches.
    
    \item We propose Co-GRPO, a unified framework that formulates the MDM model and its inference schedule as cooperating policies within a single MDP. By leveraging trajectory-level policy gradients, Co-GRPO enables cooperative optimization of both components without the computational burden of backpropagating through multi-step generation.
    
    \item Through extensive experiments, we demonstrate that Co-GRPO substantially outperforms Naive GRPO method on reward model-based benchmarks including \Imagereward \ and \Hps, while exhibiting strong zero-shot generalization across diverse text-to-image benchmarks including GenEval and DPG-Bench.
\end{itemize}

\section{Related Work}
\label{sec:related_work}
\subsection{Image Generation Models}

\noindent \textbf{Diffusion models} function by progressively refining an image from random Gaussian noise through a multi-step denoising process, with Stable Diffusion \cite{ho2020ddpm} as an early contribution and subsequent improvements in controllability and resolution \cite{rombach2022ldm, podell2023sdxl, luo2023latent, liu2024playground}. Flow matching models \cite{lipman2022flow1, liu2022flow2}, inspired by diffusion models, generate data by learning a continuous vector field that directly transforms a simple noise distribution into the target data distribution. Transformer architectures \cite{saharia2022imagen, peebles2023dit, chen2024pixart, esser2024sd3, cai2025hidream} have recently been integrated into these frameworks and have shown strong potential for performance and scalability.


\noindent \textbf{Autoregressive (AR) models} treat image generation as a next-token prediction task. VQ-VAE \cite{razavi2019vqvae} enabled the compression of images into a sequence of discrete tokens and laid the basis for a series of transformer-based AR approaches \cite{esser2021ar1vqgan, chen2018ar2pixelsnail, lee2022ar3, parmar2018ar4, team2024chameleon}. Recent studies proposed the unification of language and visual modalities, extending text comprehension and reasoning abilities to text-to-image synthesis \cite{wu2025ar6janus, deng2025ar7bagel, fang2025ar8got}. However, AR models suffer from a major computational bottleneck during high-resolution generation~\cite{cui2025emu3, wang2025simplear} and a discount in performance due to the unidirectional prior introduced by causal attention.


\noindent \textbf{Masked diffusion models (MDMs)} frame image synthesis as a mask prediction problem where all the masked visual tokens are decoded in a small, fixed number of steps, enabling remarkably fast inference. MaskGIT \cite{chang2022maskgit} pioneered this masked image modeling approach and demonstrated high fidelity and diversity. Subsequent works \cite{ni2024autonat, 10.5555/3737916.3740786, ni2024adanat, 11223107} have investigated various approaches to improve MaskGIT's generation efficiency. This concept was later extended to text-to-image (T2I) generation \cite{li2023mage, chang2023muse, bai2024meissonic} and to the unification of understanding and generation \cite{li2024mar, xie2024showo}. 
Given their recent advances, MDMs represent a highly promising direction for research and deserve further exploration.

\subsection{RL in Text-to-Image Generation}
Reinforcement learning (RL) has proven effective across diverse domains, from mathematical reasoning and code generation~\cite{lee2023rlaif,shao2024grpo} to visual perception~\cite{wang2025emulating,chen2025visrl}, demonstrating its versatility in optimizing non-differentiable decision processes. 
In image generation, RL is a powerful paradigm for aligning outputs with human preferences. Policy gradient methods such as PPO \cite{schulman2017ppo} serve as a foundational class of algorithms in this domain. DPOK \cite{fan2023dpok} and DDPO \cite{black2023ddpo} adapted them to diffusion models, enabling fine-tuning based on downstream reward functions without requiring differentiable metrics. DPO and its variants \cite{rafailov2023dpo, wallace2024diffudpo, yuan2024selfdpo, liang2024stepdpo, zhang2025diffudpo2} reformulated the objective to learn directly from preference data. Most recently, GRPO-based methods~\cite{shao2024grpo} have emerged as a promising direction. Previous work\cite{sun2025grpodiffu} demonstrated GRPO's effectiveness on diffusion models, while Flow-GRPO \cite{liu2025flowgrpo} and DanceGRPO \cite{xue2025dancegrpo} adapted the framework to flow matching models. MMaDA \cite{yang2025mmada} and Mask-GRPO \cite{luo2025maskgrpo} further extended it to MDMs, proposing preliminary techniques to estimate the transition probabilities. Our work builds upon this approach and seeks to further enhance the model's performance.



\section{Preliminaries}
\label{sec:preliminaries}
\subsection{Masked Diffusion Models for T2I Generation}

Let \(\textbf{V}\in\{1,\dots ,V\}^{N}\) denote a sequence of discrete image tokens and \(c\) the conditional prompt. A Masked Diffusion Model (MDM) aims to learn the data distribution \(p_{\mathsf{data}}(\textbf{V}|c)\) by progressively refining an initially fully-masked sequence, denoted as \((\texttt{[M]},\dots,\texttt{[M]})\), through a series of denoising steps. At each denoising step \(t\), the model predicts a conditional distribution over the next state:
\begin{equation}
    \textbf{V}^{(t)}\sim p_{\theta,t}\!\left(\,\cdot\mid \textbf{V}^{(t-1)},c\right),
\end{equation}
where \(\theta\) represents the model parameters. This distribution is estimated via a two-stage procedure:
\begin{enumerate}
    \item \textbf{Sampling step.} 
    For each position \(i\), a new token \(V_{i}^{(t)}\) is sampled or retained:
   \begin{equation}
       V_{i}^{(t)}\!\!= \!\!
   \begin{cases}
   \sim \! p_{\theta,\tau_{s}(t), s(t)}\!\left(V_{i}\mid \textbf{V}^{(t-1)},c\right),&\!\!\!\!\!\text{if }V_{i}^{(t-1)}\!\!=\!\!\texttt{[M]},\\
   V_{i}^{(t-1)},&\!\!\!\!\!\text{otherwise},
   \end{cases}
   \end{equation}
    where the \(\tau_s(t)\) and \(s(t)\) are the sampling temperature and the classifier-free guidance scale, respectively. Simultaneously, a confidence score \(C_{i}^{(t)}\) is assigned to the newly sampled tokens:
\begin{equation}
    C_{i}^{(t)}\!\!=\!\!
   \begin{cases}
   \log p\!\left(V_{i}\!=\!\hat{V}_{i}^{(t)}\!\mid\! \textbf{V}^{(t-1)}\!,c\right)\!,&\!\!\!\!\text{if }V_{i}^{(t-1)}\!\!=\!\!\texttt{[M]},\\
   +\infty,&\!\!\!\!\text{otherwise}.
   \end{cases}
\end{equation}
   
    \item \textbf{Remask step.} Let \(\tau_{r}(t)\) and \(r(t)\) be the given re-mask temperature and re-mask ratio. The re-masking distribution is defined as:
\begin{equation}
   \hat{p}_{\tau_{r}(t)}\propto\operatorname{Softmax}\!\left(C^{(t)}/\tau_{r}(t)\right).
\end{equation}
   We sample a set of indices \(U^{(t)}\subseteq\{1,\dots ,N\}\) containing \(\lceil r(t)\,N\rceil\) tokens according to \(\hat{p}_{\tau_{r}(t)}\). Selected positions are re-masked to $\texttt{[M]}$ for subsequent refinement:
\begin{equation}
   V_{i}^{(t)}\gets\texttt{[M]}\quad \forall i\in U^{(t)}.
\end{equation}
\end{enumerate}

For a comprehensive description of the sampling and re-mask procedures, please refer to~\cite{ni2024autonat}.

\subsection{Group Relative Policy Optimization (GRPO) for MDMs}
Reinforcement Learning (RL) is formally described by a discounted Markov Decision Process (MDP) defined by the tuple $(\mathcal{S}, \mathcal{A}, \mathcal{P}, \rho_0, R, \gamma)$, where $\mathcal{S}$ is the state space, $\mathcal{A}$ is the action space, $\mathcal{P}(s'|s,a)$ is the transition kernel, $\rho_0(s)$ is the initial-state distribution, $R:\mathcal{S}\times\mathcal{A}\to\mathbb{R}$ is the reward function, and $\gamma\in[0,1)$ is the discount factor. The objective in the RL framework is to find an optimal policy $\pi^{*}$ that maximizes the expected discounted return:
\begin{equation}
\pi^{*}\in\arg\max_{\pi}\mathbb{E}_{\tau\sim\pi}\!\left[\sum_{t=0}^{\infty}\gamma^{t}R(s_t,a_t)\right],    
\end{equation}
where the expectation is taken over the trajectory distribution $\tau=(s_0,a_0,s_1,a_1,\dots)$ induced by the policy $\pi$ and the environment dynamics.

We formulate the MDM generative process into a finite-horizon MDP with horizon $T$. The components are formally defined as follows:
\begin{equation}
\label{eq:naive-grpo-formulation}
\boxed{
    \begin{split}
        &s_t \triangleq (\mathbf{V}^{(t)}, c),\quad a_t \triangleq \mathbf{V}^{(t+1)}, \\ &p(s_{t+1}|s_t,a_t) \triangleq (\delta_c,\delta_{\mathbf{V}^{(t+1)}}), \\
        &p(a_t\mid s_t) \triangleq \pi_{\text{model},t} = p_{\theta,t}(\mathbf{V}^{(t+1)}\mid\mathbf{V}^{(t)},c).
    \end{split}}
\end{equation}
The model policy $\pi_{\text{model},t}$ is the conditional token prediction policy. $\delta_{x}(\cdot)$ denotes the Dirac delta function centered at $x$, which implies that the state transition is deterministic given the action $a_t$. The reward $R(s_t, a_t)$ is sparse, provided only at the terminal step $T-1$, and evaluates the quality of the completed sequence $\mathbf{V}^{(T)}$ against the condition $c$. 

For GRPO training, accurate estimation of the likelihood of $\pi_{\text{model},t}$ is crucial. Since the action $a_t$ only involves sampling new tokens at the currently masked positions, prior work~\cite{huang2025reinforcing,luo2025maskgrpo} approximates the single-step log-likelihood based on the joint probability distribution of the tokens decoded in this step. Let $I_{t}^{\mathsf{fill}} = \{i \mid V_i^{(t)} = \texttt{[M]} \text{ and } V_i^{(t+1)} \neq \texttt{[M]}\}$ be the set of indices where a token was sampled. The policy log-likelihood is approximated as the product of independent probabilities:

\begin{equation}
   \pi_{\text{model},t} \approx \prod_{i \in I_{t}^{\mathsf{fill}}} p_{\theta}\left(V_i^{(t+1)} \mid \mathbf{V}^{(t)}, c\right). 
   \label{eq:pi-model}
\end{equation}
The standard GRPO objective function is given by:
\begin{equation}
    \mathcal{L}_{\theta}=-\frac{1}{G}\frac{1}{T}\sum_{g=1}^{G}\sum_{t=0}^{T-1}[\min \Bigl(r^{g}_{t}(\theta)A^{g}_{t},\;\operatorname{clip}\bigl(r^{g}_{t}(\theta), 1-\epsilon,1+\epsilon\bigr)A^{g}_{t}\Bigr)+\beta\,\mathbb{D}_{\mathrm{KL}}\!\left(\pi_{\theta}\middle\|\pi_{\mathrm{ref}}\right)],
\end{equation}
where $r^{g}_{t}(\theta)$ is the probability ratio, defined as:
\begin{align}
    r^{g}_{t}(\theta)=&\frac{\pi_{\theta,t}(a^{g}_t|s^{g}_t)}{\pi_{\theta_\mathrm{old},t}(a^{g}_t|s^{g}_t)} \\
    =& \prod_{i \in I_{t}^{\mathsf{fill}}} \frac{p_{\theta}\left(V_{g,i}^{(t+1)} \mid \mathbf{V}_{g}^{(t)}, c\right)}{p_{\theta_\mathrm{old}}\left(V_{g,i}^{(t+1)} \mid \mathbf{V}_{g}^{(t)}, c\right)}.
\end{align}
Here, $\pi_{\theta_\text{old},t}$ is the policy used for trajectory collection. $\pi_{\text{ref}}$ is the reference model used for KL regularization, and $A^g_{t}$ is the group advantage estimated from the normalized reward.

\section{Method}
\label{sec:method}
\subsection{From Naive GRPO to Co-GRPO}
In this section, we introduce our Co-Optimized Group Relative Policy Optimization (Co-GRPO) framework, which extends the standard Markov Decision Process (MDP) for Masked Diffusion Models (MDMs). The core idea of Co-GRPO is to treat the inference schedule—specifically, sampling temperature $\tau_s$, classifier-free guidance scale $s$, re-mask temperature $\tau_r$, and re-mask ratio $r$ (denoted collectively as $\mathcal{A}$)—not as fixed hyperparameters, but as trainable actions selected by the agent at each denoising step.

Formally, we formulate this unified, finite-horizon MDP with the following components:
\begin{equation}
\label{eq:our-grpo-formulation}
\boxed{
    \begin{split}
        &s_t \triangleq (\mathbf{V}^{(t)}, c),\quad a_t \triangleq \left(\mathbf{V}^{(t+1)},\textcolor{RoyalBlue}{\mathcal{A}_{t+1}}\right), \\ &p(s_{t+1}|s_t,a_t) \triangleq (\delta_c,\delta_{\mathbf{V}^{(t+1)}}), \\
        &p(a_t\mid s_t) \triangleq p_{\theta,\textcolor{RoyalBlue}{\phi},t}(\mathbf{V}^{(t+1)},\textcolor{RoyalBlue}{\mathcal{A}_{t+1}}\mid\mathbf{V}^{(t)}, c).
    \end{split}}
\end{equation}

Here, the state $s_t$ consists of the current tokens $\mathbf{V}^{(t)}$ and conditional prompt $c$. The action $a_t$ is now a composite tuple containing both the next visual tokens $\mathbf{V}^{(t+1)}$ and the next inference schedule $\mathcal{A}_{t+1}$. Critically, the joint policy $p(a_t|s_t)$ is parameterized by both the MDM $\theta$ and the scheduling policy $\phi$.
This perspective expands the original action space of the MDP to co-optimize both the visual tokens and the schedule itself.

This formulation modularly encapsulates the conventional Naive-GRPO.  Specifically, if the policy component for the schedule is set to a \emph{fixed}, pre-defined function, i.e., $\mathcal{A}_{t} \equiv \mathcal{A}_{\text{fixed}}(t)$, our Co-GRPO framework precisely reduces to the Naive-GRPO formulation (\cref{eq:naive-grpo-formulation}).

However, this fixed-schedule assumption is a significant limitation. To illustrate the sensitivity of the schedule, we conducted a simple experiment (\cref{tab:simple-exp}). We tested a cosine schedule for the mask ratio, $r_t=\cos{(\frac{t+1}{T})^\gamma}$, varying only the exponent $\gamma$. The results show that minor changes in the schedule lead to significant variations in evaluation metrics.


\begin{figure*}[t]        
  \centering
  \includegraphics[width=0.9\linewidth]{fig/model.pdf}
  \caption{\textbf{Overview of our proposed Co-GRPO.} During the trajectory collection phase, both the sampled visual tokens ($\mathbf{V}$) and their associated inference schedule ($\mathcal{A}$) are collected at each step. These trajectories are evaluated by the reward model, and the resulting scores are aggregated and normalized at the group level to compute individual advantages. In the subsequent policy optimization phase, the joint policy is explicitly factorized into a \emph{model policy} $\pi_{\theta}$ and a \emph{scheduling policy} $\pi_{\phi}$. By estimating their respective likelihoods and applying an alternating optimization strategy, our approach enables the cooperative refinement of both policies toward improved generation quality.}
  \label{fig:model}
\end{figure*}
Motivated by this finding, we move beyond fixed schedule and formalize $\mathcal{A}$ as a trainable action. This leads to a factorization of the joint policy $p(a_t|s_t)$ from \cref{eq:our-grpo-formulation}:
\begin{equation}
\begin{split}
    p(a_t|s_t) &\triangleq p_{\theta,\textcolor{RoyalBlue}{\phi},t}(\mathbf{V}^{(t+1)},\textcolor{RoyalBlue}{\mathcal{A}_{t+1}}\mid\mathbf{V}^{(t)}, c) \\
&=  p_{\theta,t}(\mathbf{V}^{(t+1)} \!\mid\! \textcolor{RoyalBlue}{\mathcal{A}_{t+1}}; \mathbf{V}^{(t)}, c) \!\cdot\! p_{\textcolor{RoyalBlue}{\phi},t}(  \textcolor{RoyalBlue}{\mathcal{A}_{t+1}} \!\mid\! \mathbf{V}^{(t)}, c)  \\
&= \pi_{\text{model},t} \cdot \pi_{\text{schedule},t} \\
\end{split}
\label{eq:joint_policy}
\end{equation}

Here, $\pi_{\text{model},t}$ is the model policy, analogous to the policy in Naive-GRPO, which now generates $\mathbf{V}^{(t+1)}$ conditioned on the dynamically chosen schedule $\mathcal{A}_{t+1}$. The second term, $\pi_{\text{schedule},t}$, is the new scheduling policy that learns to select $\mathcal{A}_{t+1}$ based on the current state $s_t$.

We model the scheduling policy $\pi_{\text{schedule},t}$ as a multivariate Gaussian distribution, whose mean is predicted by a network, to parameterize the continuous components of $\mathcal{A}$:
\begin{equation}
    \pi_{\text{schedule},t} = p_{\phi, t} \left( \mathcal{A}_{t+1}\mid\textbf{V}^{(t)},c \right)
    \sim \mathcal{N}(\eta_\phi\left(s_t\right), \sigma \mathbf{I}),
\end{equation}
where $\eta_\phi(s_t)$ is a network parameterized by $\phi$ that predicts the mean of the distribution, and $\sigma$ is a fixed hyperparameter controlling the policy's exploration variance.

This extended MDP exposes both the denoising network ($\theta$) and the scheduling network ($\phi$) to the same reinforcement signal. Consequently, the Co-GRPO framework optimizes a unified clipped surrogate objective:

\begin{equation}
\label{eq:co-train-obj}
    \mathcal{L}_{\theta,\phi}=-\frac{1}{G}\frac{1}{T}\sum_{g=1}^{G}\sum_{t=0}^{T-1}[\min\!\Bigl({r^g_{t}}(\theta,\phi)A^g_{t},\;\operatorname{clip}\!\bigl({r^g_{t}}(\theta,\phi), \\ 
    1-\epsilon,1+\epsilon\bigr)A^g_{t}\Bigr)+\beta\,\mathbb{D}_{\mathrm{KL}}\!\left(\pi_{\theta,\phi}\middle\|\pi_{\mathrm{ref}}\right)],
\end{equation}
where the probability ratio $r^g_{t}(\theta, \phi)$ is based on joint policy:
\begin{align}
    r^{g}_{t}(\theta,\phi)=&\frac{\pi_{\theta,\phi,t}(a^{g}_t|s^{g}_t)}{\pi_{\theta_\text{old},\phi_\text{old}, t}(a^{g}_t|s^{g}_t)}.
\end{align}


The objective explicitly allows gradients to flow to both model $\theta$ and schedule $\phi$. This enables the denoising network and its own inference schedule to \emph{co-adapt} simultaneously, optimizing for the expected return.

\subsection{Alternating Co-Optimization Strategy}
\label{subsec:alternative-co-optimize}

\begin{wraptable}{r}{0.48\columnwidth} 
    \centering
    \footnotesize
    \vspace{-10pt}
    \begin{tabular}{cccc}
    \toprule
       Steps  & $\gamma = 1.0$ & $\gamma = 1.5$ & $\gamma = 2.0$\\
       \midrule
        16 & \cellcolor{gray!20}0.660 & 0.872 & \textbf{0.865} \\
        48 & \cellcolor{gray!20}0.918 & \textbf{0.957} & 0.937 \\
    \bottomrule
    \end{tabular}
    \vspace{-5pt}
    \caption{\textbf{Preliminary ablation on the impact of the inference schedule.} We perturb the cosine masking schedule by introducing a variance factor $\gamma$: $r_t = \cos{\big(\pi(t+1)/2T\big)^\gamma}$. Results on the ImageReward~\cite{xu2023imagereward} demonstrate that scheduling parameters critically affect MDM generation quality, exposing the limitations of fixed schedules in Naive GRPO. Default setting is marked in \colorbox{gray!20}{gray}.}
    \vspace{-15pt}
    \label{tab:simple-exp}
\end{wraptable}

As depicted in \cref{fig:model}, our Co-GRPO (Co-Optimized Group Relative Policy Optimization) framework aims to jointly train the denoising model ($\theta$) and the scheduling policy ($\phi$). The foundation for this joint training lies in the factorization of the joint policy likelihood (\cref{eq:joint_policy}), which estimates the trajectory based on both the model's token generation and the schedule's action selection.

A naive, simultaneous optimization, however, presents a significant challenge. Specifically, the denoising model ($\theta$) contains vastly more parameters than the scheduling policy ($\phi$). Yet, as demonstrated in our preliminary experiment (\cref{tab:simple-exp}), the low-dimensional schedule $\mathcal{A}$ governed by $\phi$ has a disproportionately large impact on generation quality. This asymmetry—a small network wielding significant influence—creates an unstable training dynamic when both $\theta$ and $\phi$ are updated concurrently, which leads to suboptimal convergence (see \cref{subtab:training_strategy}).

To address this challenge, we propose to leverage the separability inherent in our policy factorization (\cref{eq:joint_policy}). We revisit the joint probability ratio $r^g_{t}$ (denoted as $r_t$ in the subsequent derivation for notational simplicity) and decompose it according to our factorization:
\begin{align}
\label{eq:ratio_factorization}
    r_{t}(\theta,\phi) = \frac{\pi_{\theta,\phi,t}(a_t|s_t)}{\pi_{\theta_\mathrm{old},\phi_\mathrm{old},t}(a_t|s_t)} 
    = \frac{\pi_{\text{model},t} \cdot \pi_{\text{schedule},t}}{\pi_{\text{old model},t} \cdot \pi_{\text{old schedule},t}}
    = \frac{\pi_{\text{model},t}}{\pi_{\text{old model},t} } \cdot\frac{\pi_{\text{schedule},t}}{\pi_{\text{old schedule},t}}
\end{align}
where
\begin{align}
    \frac{\pi_{\text{model},t}}{\pi_{\text{old model},t} } &= \frac{p_{\theta,t}(\mathbf{V}^{(t+1)} \mid \textcolor{black}{\mathcal{A}_{t+1}}; \mathbf{V}^{(t)}, c)}{p_{\theta_{\mathrm{old}},t}(\mathbf{V}^{(t+1)} \mid \textcolor{black}{\mathcal{A}_{t+1}}; \mathbf{V}^{(t)}, c)} \\
    &= \prod_{i \in I_{t}^{\mathsf{fill}}}  \frac{p_{{\theta}, t}\left(V_i^{(t+1)} \mid  \textcolor{black}{\mathcal{A}_{t+1}} ; \mathbf{V}^{(t)}, c\right)}{p_{{\theta_{\mathrm{old}}}, t}\left(V_i^{(t+1)} \mid  \textcolor{black}{\mathcal{A}_{t+1}} ; \mathbf{V}^{(t)}, c\right)},
\end{align}
is independent of the scheduling control parameters $\phi$ once $\textcolor{black}{\mathcal{A}_{t+1}}$ is determined, and 
\begin{equation}
    \begin{split}
        \frac{\pi_{\text{schedule},t}}{\pi_{\text{old schedule},t}} &= \frac{p_{\phi,t}(  \textcolor{black}{\mathcal{A}_{t+1}} \mid \mathbf{V}^{(t)}, c)}{p_{\phi_{\mathrm{old}},t}(  \textcolor{black}{\mathcal{A}_{t+1}} \mid \mathbf{V}^{(t)}, c)}
    \end{split},
\end{equation}
is also independent of the model parameters $\theta$.  In other words, the two sets of trainable parameters $\theta$ and $\phi$ are highly separable in our optimization objective.

Driven by this observation, we propose an alternative optimization strategy that decouples the training into distinct phases. 
The optimization alternates between $N_m$ iterations of model parameter updates and $N_s$ iterations of schedule updates. In this scheme, we modify the probability ratio $r_{t}$ used in the Co-GRPO objective (\cref{eq:co-train-obj}) based on the current phase. Formally, within a single update cycle ($n=1, \dots, N_m+N_s$), the ratio is defined as:
\begin{equation}
\label{eq:ours-prob-ratio}
r_{t}(\theta,\phi) = 
    \begin{cases}
        \frac{\prod_{i \in I_{t}^{\mathsf{fill}}}  p_{{\theta}, t}\left(V_i^{(t+1)} \mid  {\mathcal{A}_{t+1}} ; \mathbf{V}^{(t)}, c\right)}{\prod_{i \in I_{t}^{\mathsf{fill}}}  p_{\theta_\text{old}, t}\left(V_i^{(t+1)} \mid  {\mathcal{A}_{t+1}} ; \mathbf{V}^{(t)}, c\right)} & n < N_m \\
        \frac{p_{\phi,t}(  {\mathcal{A}_{t+1}} \mid \mathbf{V}^{(t)}, c)}{p_{\phi_\text{old},t}(  {\mathcal{A}_{t+1}} \mid \mathbf{V}^{(t)}, c)} & \text{otherwise}.
    \end{cases}
\end{equation}
This approach ensures that each component is optimized with respect to a stable counterpart, significantly improving convergence behavior and overall performance. The ablation study in \cref{subtab:training_strategy} also proves this point.

\section{Experiments}
\label{sec:experiments}

\begin{table}[!t]
    \centering
    \footnotesize
    \hspace{-10pt}
    \begin{tabular}{lccccccc}
    \toprule
\multicolumn{1}{c}{\multirow{2}{*}{\textbf{Model}}} & \multicolumn{1}{c}{\multirow{2}{*}{\textbf{Params}}} & \multicolumn{1}{c}{\multirow{2}{*}{\textbf{\Imagereward}}} & \multicolumn{5}{c}{\textbf{HPS v2.0}} \\
    \cline{4-8}
    \addlinespace[0.3em] 
    \multicolumn{1}{c}{} & \multicolumn{1}{c}{} & \multicolumn{1}{c}{} & {\textbf{Animation}} & {\textbf{Concept-art}} & {\textbf{Painting}} & {\textbf{Photo}} & {\textbf{Averaged}} \\
    \midrule
    \multicolumn{8}{c}{\textbf{Diffusion-based Models}}
    \\
    \addlinespace[0.3em] 
        SD v1.4~\cite{rombach2022ldm} & 0.9B & 0.087 & 27.26 & 26.61 & 26.66 & 27.27 & 26.95 \\
        SD v2.0~\cite{rombach2022ldm} & 0.9B & 0.174 & 27.48 & 26.89 & 26.86 & 27.46 & 27.17 \\
        Dreamlike Photoreal 2.0~\cite{dreamlike} & 0.9B & 0.168 & 28.24 & 27.60 & 27.59 & 27.99 & 27.86\\
        DeepFloyd-XL~\cite{deepfloyd} & 5.5B & 0.453 & 27.64 & 26.83 & 26.86 & 27.75 & 27.27\\
        SDXL Base 1.0~\cite{podell2023sdxl} & 3.5B & 0.790 & 28.88 & 27.88 & 27.92 & 28.31 & 28.25 \\
        SDXL Refiner 1.0~\cite{podell2023sdxl} & 6.6B & 0.886 & 28.93 & 27.89 & 27.90 & 28.38 & 28.27\\
        DALL-E 3~\cite{betker2023dalle3} & {-} & 1.094 & 29.09 & 28.07 & 28.15 & 28.41 & 28.43 \\
        Majicmix Realistic v7~\cite{merjic2023majicmix} & 0.9B & 0.126 & 27.88 & 27.19 & 27.22 & 27.64 & 27.48\\
        Pixart-$\alpha$~\cite{chen2024pixart} & 0.6B & 1.115 & 29.30	& 28.57 &	28.55	& 28.99	& 28.85 \\
        Deliberate~\cite{deliberate} & 1.5B & 0.285 & 28.13 & 27.46 & 27.45 & 27.62 & 27.67\\
        Realistic Vision~\cite{SG_1612222024realvis} & 1.1B & 1.088 & 28.22 & 27.53 & 27.56 & 27.75 & 27.77\\
        OmniGen~\cite{xiao2025omnigen} & 3.8B & 1.055 & 29.22 & 28.25 & 28.43 & 28.89 & 28.70\\
        \midrule
        \multicolumn{8}{c}{\textbf{Token-based Models}}
        \\
        \addlinespace[0.3em] 
        Show-o~\cite{xie2024showo} & 1.3B & 1.028 & 28.90 & 28.34 & 28.40 & 28.47 & 28.53\\
        BLIP3o-NEXT~\cite{chen2025blip3onext} & 3.0B & 0.926 & 28.32 & 27.43 & 27.53 & 28.44 & 27.93 \\
        MaskGen-XL~\cite{kim2025maskgen} & 1.1B & 0.797 & 28.25 & 27.98 & 27.83 & 27.78 & 27.96\\
        Meissonic~\cite{bai2024meissonic} & 1.0B & 0.942 & 29.57 & 28.58 & 28.72 & 28.45 & 28.83\\
        Meissonic\,+\,Co-GRPO & 1.0B & {\textbf{1.122\rlap{$_{\text{\scriptsize\textcolor{ForestGreen}{(+0.18)}}}$}}} & {\textbf{29.70\rlap{$_{\text{\scriptsize\textcolor{ForestGreen}{(+0.13)}}}$}}} & {\textbf{29.62\rlap{$_{\text{\scriptsize\textcolor{ForestGreen}{(+1.04)}}}$}}} & {\textbf{29.12\rlap{$_{\text{\scriptsize\textcolor{ForestGreen}{(+0.40)}}}$}}} & {\textbf{29.03\rlap{$_{\text{\scriptsize\textcolor{ForestGreen}{(+0.58)}}}$}}} & {\textbf{29.37\rlap{$_{\text{\scriptsize\textcolor{ForestGreen}{(+0.54)}}}$}}}\\
        
        \bottomrule
    \end{tabular}
    \caption{Quantitative results on reward model based benchmarks \ImagerewardCite \ and \HpsCite.}
    \label{tab:main-reward}
    \vspace{-5pt}
\end{table}
\subsection{Implementation Details}

\subsubsection{Training Strategy}
We trained our Co-GRPO model utilizing a composite reward signal derived from two model-based reward: \ImagerewardCite \ and \HpsCite. The advantage calculation employed a weighted linear combination of these two components, with both models contributing equally (a weight of $0.5$ for each component) to the total advantage.

The base text-to-image model employed is Meissonic~\cite{bai2024meissonic}, a high-performance Masked Diffusion Model. Consistent with the default configuration of Meissonic, the number of inference steps was fixed at 48 throughout the reinforcement learning training process. The Kullback-Leibler ($\mathrm{KL}$) divergence regularization coefficient $\beta$ was set to $0$, consistent with prior reinforcement learning studies applied to MDMs~\cite{luo2025maskgrpo,huang2025reinforcing}. Further detailed hyperparameter configurations and the specific network architectures pertaining to the alternative co-optimization strategy are provided in the \cref{sec:supp-implementation}.

\subsubsection{Evaluation Details}
During evaluation, the model's inference step count is fixed at 48 to maintain consistency with the training setup. Performance under varying step counts is further investigated and presented in \cref{subtab:step_ability}.
We report the model's performance on the \Imagereward \ and \Hps \ rewards. The former consists of 500 prompts, and the latter contains approximately 3,000 prompts. The \Hps \ rewards are further categorized into four distinct image categories: Animation, Concept-art, Painting, and Photo. Additionally, to demonstrate the model's generalization capability, we test its performance on two established general text-to-image benchmarks \GenevalCite \ and \DpgCite.

\vspace{10pt}

\subsection{Main Results}

\begin{wraptable}{r}{0.48\columnwidth} 
\centering
\footnotesize
\vspace{-10pt}
\begin{adjustbox}{width=\linewidth}
\begin{tabular}{lccc}
\toprule
Model & Params & \Geneval &  \Dpg \\
\midrule
DALL-E mini~\cite{dallemini} & 0.4B & 0.23 & - \\
DALL-E 2~\cite{ramesh2022dalle2} & 3.5B & 0.52	& - \\
SD v1.5~\cite{rombach2022ldm} & 0.9B & 0.43	& 63.18 \\
SD v2.1~\cite{rombach2022ldm} & 0.9B & 0.50	& 68.09 \\
LlamaGen~\cite{sun2024llamagen} & 0.8B & 0.32 & 65.16 \\
Chameleon~\cite{team2024chameleon} & 7.0B & 0.39 & - \\
\midrule
Meissonic~\cite{bai2024meissonic} & 1.0B & 0.47 & 64.57 \\
Meissonic\,+\,Co-GRPO & 1.0B & \textbf{0.55\rlap{$_{\text{\scriptsize\textcolor{ForestGreen}{(+0.08)}}}$}} & \textbf{70.10\rlap{$_{\text{\scriptsize\textcolor{ForestGreen}{\textbf{(+6.53)}}}}$}} \\
\bottomrule
\end{tabular}
\end{adjustbox}
\vspace{-5pt}
\caption{\textbf{Quantitative results on general prompt-adherence benchmarks.} Our model is trained under the same configuration as the main experiment using the ImageReward and HPSv2 reward models, and evaluated in a zero-shot setting on the \GenevalCite \ and \DpgCite \ benchmarks without relying on external distilled data or ground-truth detectors.}
\label{tab:main-zeroshot}
\vspace{-15pt}
\end{wraptable}


The results presented in \cref{tab:main-reward} demonstrate that our method substantially improves the \Imagereward \ and \Hps \ rewards while introducing only a marginal increase in learnable parameters. Specifically, Co-GRPO training delivers notable performance gains: \Imagereward \ increases by $0.18$ and the \Hps \ reward improves by $0.54$, outperforming all reported baselines. Notably, our 1B-parameter model surpasses models with significantly larger number of parameters, underscoring the superior efficiency and effectiveness of our approach in aligning with human preferences.

Furthermore, \cref{tab:main-zeroshot} demonstrates the strong generalization capability of our model on established text-to-image evaluation benchmarks. Importantly, Co-GRPO is trained without using any prompts or reward signals from \Geneval \ or \Dpg, making this a zero-shot evaluation setting. Despite this, our method achieves substantial improvements, elevating the \Geneval \ score from $0.47$ to $0.55$ and the \Dpg \ score from $64.57$ to $70.10$. 
These gains on unseen benchmarks demonstrate that Co-GRPO learns generalizable human preference alignment that transfers effectively to diverse text-to-image generation tasks.

\subsection{Ablation Study}

\cref{tab:comprehensive_ablation} presents the comprehensive results of the ablation.

\noindent\textbf{Component of Trainable Action Space}
Results in \cref{subtab:component_ablation} demonstrate a monotonic improvement in both the \Imagereward \ and \Hps \ as more components of the action space are progressively made trainable. Optimizing only the model parameters (\ie, the $\text{Naive GRPO}$ formulation in \cref{eq:naive-grpo-formulation}) yields only marginal gains in generation quality. In contrast, introducing the scheduling policy within the $\text{Co-GRPO}$ framework leads to substantial performance improvements. Moreover, progressively incorporating additional components of the inference schedule produces steady, incremental gains, indicating that each component provides tangible gains to the overall model performance.

\noindent\textbf{Optimization Strategy}
\cref{subtab:training_strategy} compares the Alternative optimization strategy with the Joint optimization strategy, controlling for an identical total number of training iterations. While the Joint approach also improves model performance, its effect is less significant than the Alternative method. This validates the intuition discussed in \cref{subsec:alternative-co-optimize} and underscores the superiority and necessity of the Alternative training approach for optimizing. \cref{subtab:training_iterations} illustrates the impact of the number of alternating optimization cycles on training efficacy. We observe a substantial performance leap after the initial cycle, and subsequent cycles further reinforce the model's performance. In our experiments, performance largely converges after three cycles.


\begin{table}[t]
  \centering
  \footnotesize
  \vspace{-3pt}
  \begin{adjustbox}{valign=t}
  \hspace{-5pt}
  \hspace{-45pt}
  \begin{minipage}[t]{0.48\textwidth}
    \centering
    \begin{tabular}{ccccccc}
      \toprule
      \multirow{2}{*}{Model} & \multicolumn{4}{c}{Schedule} & \multirow{2}{*}{\Imagereward} & \multirow{2}{*}{\Hps}\\
      \cmidrule{2-5}
      & $r$ & $\tau_r$ & $\tau_s$ & $s$ & &  \\
      \midrule
       &  &  &  & & 0.942 & 28.83 \\
      \cmark &  &  &  & & 1.014 & 28.86 \\
      \cmark & \cmark &  &  &  & 1.037 & 28.93 \\
      \cmark & \cmark & \cmark &  &  & 1.070 & 29.09 \\
      \cmark & \cmark & \cmark & \cmark &  & 1.102 & 29.21 \\
\rowcolor{gray!20}      \cmark & \cmark & \cmark & \cmark & \cmark & \textbf{1.122} & \textbf{29.37} \\
      \bottomrule
    \end{tabular}
    \subcaption{Comparison on the component of trainable action space.}
    \label{subtab:component_ablation}
  \end{minipage}
  \end{adjustbox}
  \vspace{3pt}
  \begin{adjustbox}{valign=t}
  \hspace{-5pt}
  \begin{minipage}[t]{0.48\textwidth}
    \centering
    \begin{tabular}{llcccccc}
    \toprule
    \multicolumn{2}{c}{Steps} & \multicolumn{3}{c}{\Imagereward} & \multicolumn{3}{c}{\Hps} \\
    \cmidrule{3-5} \cmidrule{6-8}
    Test & Train & Baseline & Ours & $\Delta$ & Baseline & Ours & $\Delta$ \\
    \midrule 
    8  & 48 & 0.133 & 0.874 & 0.741 & 26.42 & 28.61 & 2.19 \\
    16 & 48 & 0.735 & 1.053 & 0.318 & 28.03 & 29.23 & 1.20 \\
    32 & 48 & 0.902 &  1.091 & 0.189 & 28.62 & 29.23 & 0.61 \\
    \rowcolor{gray!20}48 & 48 & 0.942 & \textbf{1.122} & 0.180 & 28.83 & 29.37 & 0.54 \\
    64 & 48 & 0.941 & 1.099 & 0.158 & 28.80 & \textbf{29.39} & 0.59 \\
    \bottomrule
    \end{tabular}
    \vspace{9pt}
    \subcaption{Comparison of transfer capability on different inference steps.}
    \label{subtab:step_ability}
\end{minipage}
\end{adjustbox}
    \vspace{10pt}
    \hspace{-20pt}
  \begin{adjustbox}{valign=t}
  \begin{minipage}[h]{0.38\textwidth}
    \centering
    \begin{tabular}{ccc}
      \toprule
      Cycle & \Imagereward & \Hps \\
      \midrule
      0 & 0.942 & 28.83 \\
      1 & 1.067 & 29.21 \\
      2 & 1.093 & 29.22 \\
      \rowcolor{gray!20}3 & \textbf{1.122} & \textbf{29.37} \\
      \bottomrule
    \end{tabular}
    \subcaption{Comparison on alternative training cycles.}
    \label{subtab:training_iterations}
  \end{minipage}
  \end{adjustbox}
   \hspace{-5pt}
  \begin{adjustbox}{valign=t}
  \begin{minipage}[h]{0.30\textwidth}
    \centering
    \begin{tabular}{ccc}
      \toprule
      Strategy & \Imagereward & \Hps \\
      \midrule
      Baseline & 0.942 & 28.83 \\
      Joint & 1.031 & 29.06 \\
      \rowcolor{gray!20} Alternative & \textbf{1.122} & \textbf{29.37} \\
      \bottomrule
    \end{tabular}
    \vspace{8pt}
    \subcaption{Comparison on training strategies.}
    \label{subtab:training_strategy}
  \end{minipage}
  \end{adjustbox}
  \hspace{20pt}
  \begin{adjustbox}{valign=t}
  \begin{minipage}[h]{0.30\textwidth}
    \centering
    \begin{tabular}{ccc}
      \toprule
      Method & \Mps & \Clipscore \\
      \midrule
      Baseline & 13.47 & 31.94 \\
      \rowcolor{gray!20}Co-GRPO & 1\textbf{4.05} & \textbf{32.35} \\
      $\Delta$ & 0.58 & 0.41 \\ 
      \bottomrule
    \end{tabular}
    \vspace{10pt}
    \subcaption{Comparison on transfer capability to different reward models.}
    \label{subtab:reward_ability}
  \end{minipage}
  \end{adjustbox}
  \vspace{-18pt}
  \caption{\textbf{Ablation studies.} We mark our default settings in \colorbox{gray!20}{gray}.}
  \label{tab:comprehensive_ablation}
  \vspace{-45pt}
\end{table}
\noindent\textbf{Transfer Capability Analysis} 
We analyze the transferability of our learned policy across different inference settings. In \cref{subtab:step_ability}, we evaluate a model trained for 48 steps under both fewer (8) and more (64) inference steps. The scheduling policy is transferred between different step counts using interpolation. We find that our method yields a consistent improvement over the baseline at all tested step counts, with the performance gain being especially prominent at smaller step counts. This result demonstrates the model's generalizability concerning the number of inference steps and suggests that the learned scheduling policy may be more critical for achieving high-quality generation under low-step conditions.
In \cref{subtab:reward_ability}, we evaluate the model's performance against other reward models (\MpsCite \, and \ClipscoreCite). The model shows consistent performance gains across these metrics, which demonstrates strong generalization ability across various external reward models.

\vspace{20pt}
\subsection{Visualization Results}

\begin{wrapfigure}{r}{0.48\columnwidth} 
    \footnotesize
    \vspace{-45pt}
    \centering
    \includegraphics[width=1\linewidth]{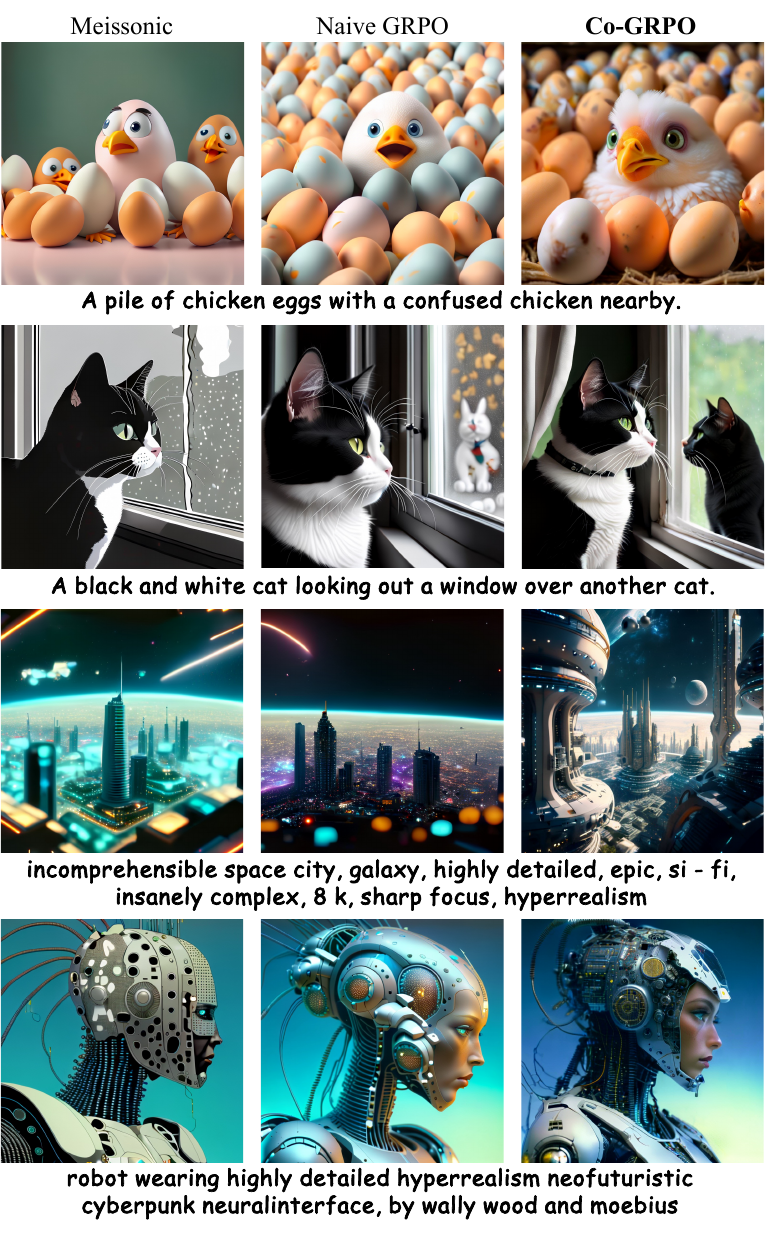}
    \vspace{-25pt}
    \caption{\textbf{Qualitative comparisons of the base model and models optimized with GRPO and Co-GRPO.} Co-GRPO generates images with superior aesthetics while better preserving fine-grained visual details compared to both the base model and GRPO.}
    \vspace{-55pt}
    \label{fig:visual-compare}
\end{wrapfigure}

\cref{fig:visual-compare} provides qualitative comparisons between images generated by our approach, the base model, and the GRPO-optimized model. Across a wide range of prompts, Co-GRPO produces outputs with consistently finer details and higher visual fidelity. Furthermore, in specific instances, such as the prompt ``A black and white cat looking out a window over another cat'', the generated image demonstrate stronger prompt adherence. This indicates that Co-GRPO improves prompt following along with better aesthetics. The enhanced visual quality and semantic accuracy collectively demonstrate the effectiveness of jointly optimizing both the model and inference schedule.

\section{Conclusion}
\vspace{10pt}
\label{sec:conclusion}

In this paper, we present Co-GRPO, a GRPO-based framework that substantially improves MDMs performance on text-to-image generation. Our work is driven by the insight that the inference schedule—an essential yet previously underexplored component of the MDM generation process—plays a pivotal role in generation performance. To address this, we reformulated the underlying MDP to incorporate the scheduling policy, establishing the basis for our unified Co-GRPO framework. Building on this formulation, we further developed the mathematical formulation for a Co-Optimization Strategy that jointly optimizes the inference schedule and model parameters. Our approach yields significant improvements and generalization results across diverse rewards and benchmarks, highlighting the importance of co-optimization during post-training.


\bibliography{main}
\bibliographystyle{icml2025}




\clearpage
\appendix
\section*{Appendix}

\renewcommand{\thesection}{\Alph{section}} 

\section{More Implementation Details}
\label{sec:supp-implementation}
\textbf{Dataset.} We conduct our experiments using a mixture of prompts from the HPDv2~\cite{wu2023human} and ImageReward~\cite{xu2023imagereward} datasets' training splits, consisting of 103,700 and 8,000 prompts respectively. We use only the text prompts from these datasets without their corresponding images.

\noindent\textbf{Model architecture.} 
Following Meissonic~\cite{bai2024meissonic}, our text encoder is CLIP-ViT-H-14 from OpenCLIP~\cite{Radford2021LearningTV}, which remains frozen during training. Our scheduling policy network consists of a depthwise convolution layer, a pointwise convolution layer, and a multi-layer perceptron (MLP). We extract visual token features from the final layer output of the transformer blocks and incorporate timestep information into the policy network using adaptive layer normalization (AdaLN)~\cite{peebles2023scalable,perez2018film}.

\noindent\textbf{Training settings.}
We optimized both policies with Adam ($\beta_1=0.9, \beta_2=0.95$).
The 1.0B-parameter model policy was trained for 300 iterations with learning rate $1\times10^{-5}$, weight-decay $0.02$, group size $G=6$, and total batch size 96.
The 9M-parameter Scheduling Policy was trained for 200 iterations with learning rate $1\times10^{-4}$, weight decay 0, $G=8$, and total batch size 256. Its lighter architecture enabled faster convergence.

\section{Comparison Across Inference Steps}
\begin{wrapfigure}{r}{0.38\columnwidth} 
    \centering
    \vspace{-10pt}
    \includegraphics[width=0.8\linewidth]{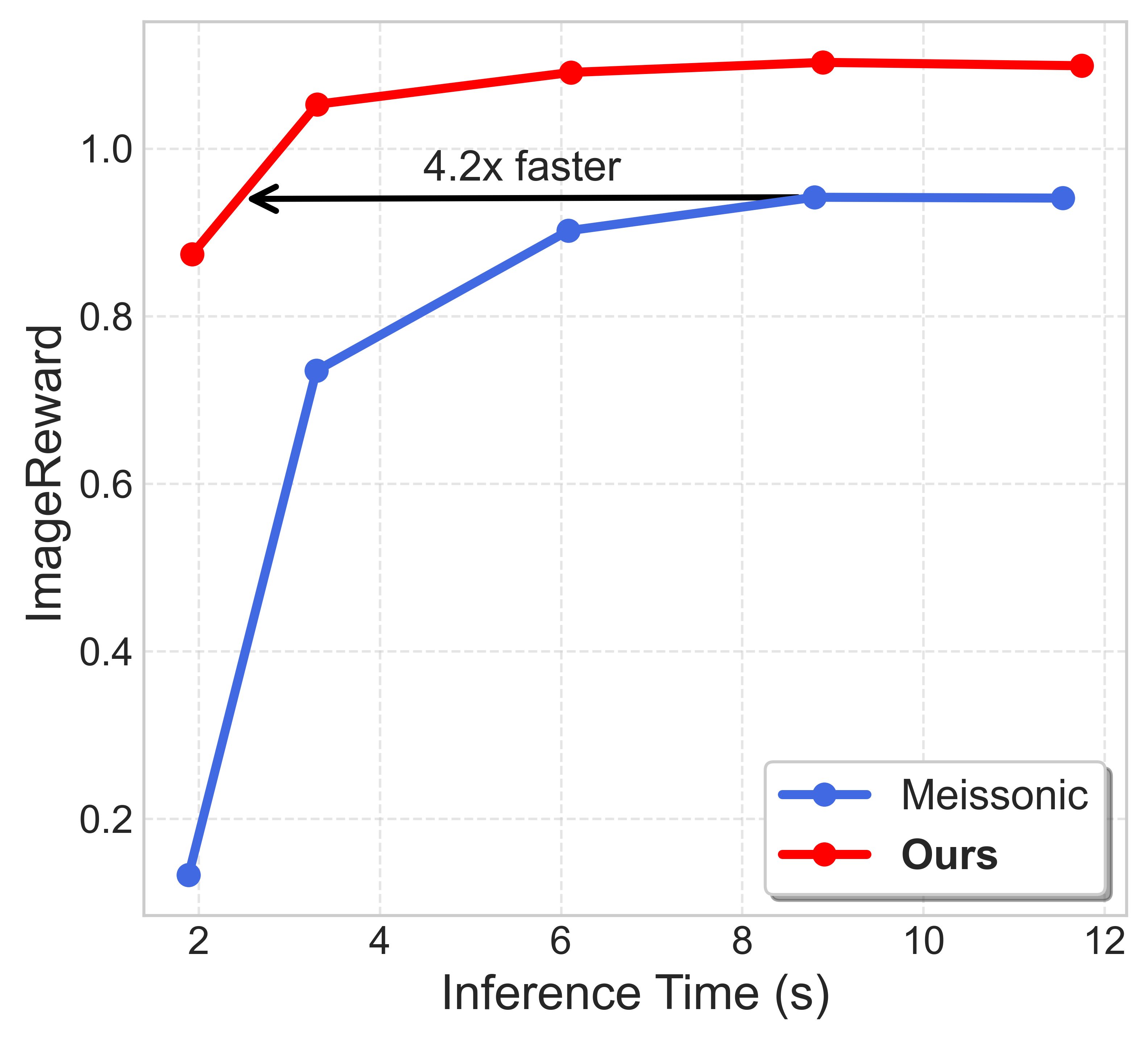}
    \caption{\textbf{ Comparison between Meissonic base model and our Co-GRPO trained model across different steps.} Our Co-GRPO improves the performance of the model over every steps, with a 4.2 times faster comparied to the base model on the same score.}
    \vspace{-10pt}
    \label{suppfig:efficiency}
\end{wrapfigure}
We conduct a comparative analysis between the Meissonic base model and our Co-GRPO method by varying the number of inference steps from 8 to 64. All results are evaluated using the same checkpoint trained with 48 steps; only the inference step count is varied at test time. We measure the overall inference time with batch size 4 on an NVIDIA A100 (40GB). Notably, the reported inference times for Co-GRPO include the computational overhead of the scheduling policy network. The number of parameters of the scheduling policy  is less than 1\% of the base model's, resulting in negligible forward pass latency that does not substantially affect overall inference time.

\begin{wraptable}{r}{0.58\columnwidth} 
\footnotesize
    \centering
    \vspace{-50pt}
    \begin{tabular}{p{0.1\linewidth}p{0.85\linewidth}}
    \toprule
        Column & \multicolumn{1}{c}{Prompts} \\
        \midrule
        1 & The woman sitting at the table looks bored. \\ 
        \midrule
        2 & A digital painting of a beautiful creature in intricate detail, centered in a low angle shot. \\ 
        \midrule
        3 & US Air Force battling against the Rebellion in a radioactive environment with detailed digital artwork. \\
        \midrule
        4 & The sea, sunset, colorful, clear water, digital illustration. \\ 
        \midrule
        5 & The image is a cinematic portrait of Walt Whitman depicted as a bodhisattva in the style of several famous artists, painted in oil on canvas or gouache with intricate details and desaturated colors. \\
        \bottomrule
    \end{tabular}
    \caption{Prompts employed in the teaser examples generation.}
    \label{supptab:teaser-prompt}
\end{wraptable}

As illustrated in \cref{suppfig:efficiency}, Co-GRPO consistently improves ImageReward scores across all step counts, demonstrating that the learned scheduling policy effectively reallocates computational resources to the most informative timesteps. Specifically, while the baseline model requires 48 steps to achieve an ImageReward of 0.94, Co-GRPO attains the same performance with fewer than 16 steps, representing a great reduction of over required inference steps.

\section{Visualization Results}
\subsection{Prompts used for Teaser Figure}
We present the prompts used for the teaser figure in \cref{supptab:teaser-prompt}. 
\subsection{More Visualization Results}
\begin{figure*}[t]
\centering
\includegraphics[width=\linewidth]{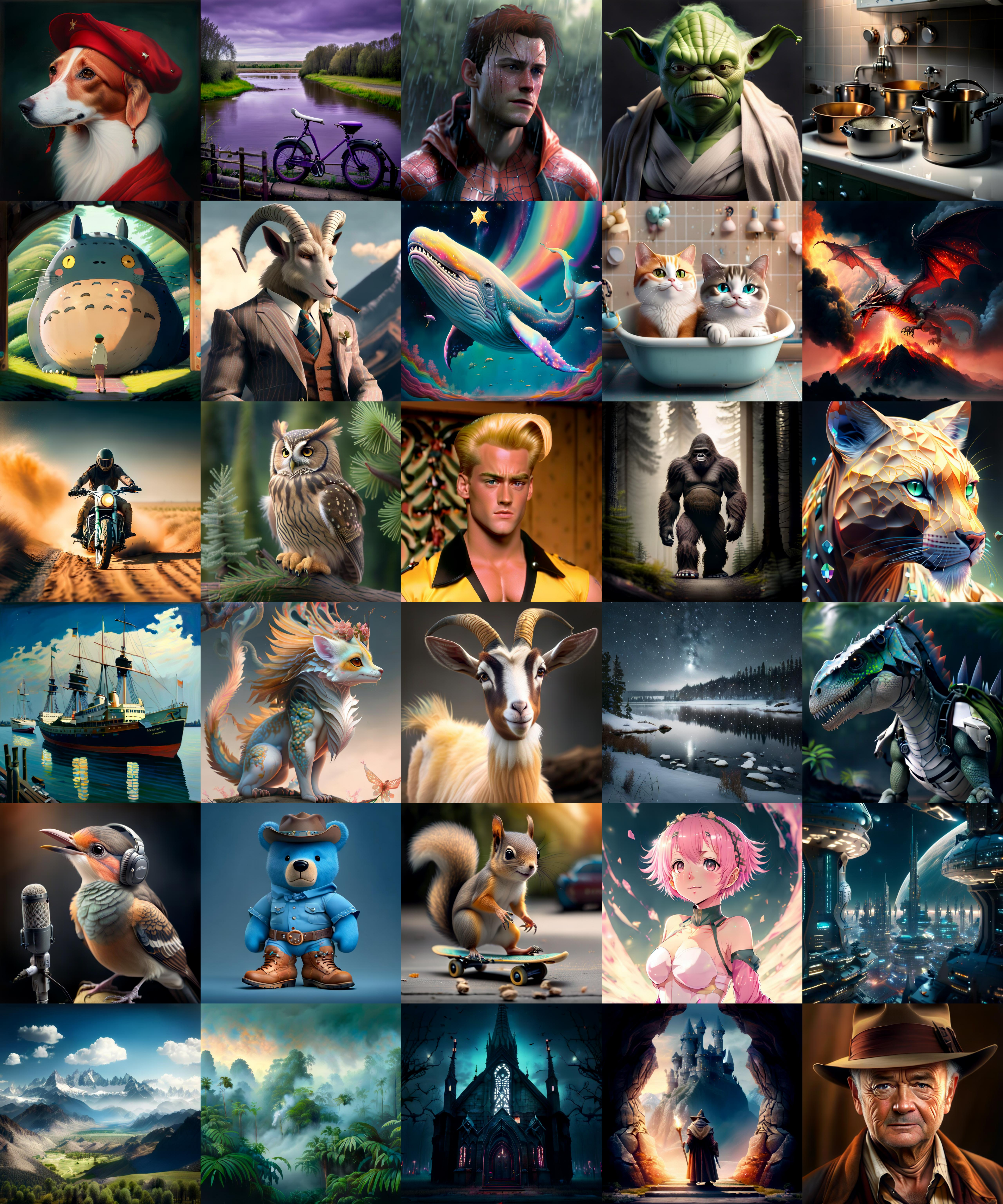}
\caption{Representative high-quality images generated by our Co-GRPO method.}
\label{suppfig:our-best}
\end{figure*}

Additional images generated by our Co-GRPO method are shown in \cref{suppfig:our-best}. All of the prompts are selected from ImageReward and HPDv2 test splits.


\end{document}